\documentclass[10pt,twocolumn,letterpaper]{article}

\usepackage{cvpr}              

\usepackage{url}            %
\usepackage{booktabs}       %
\usepackage{amsfonts}       %
\usepackage{nicefrac}       %
\usepackage{microtype}      %
\usepackage{graphicx}
\usepackage{xcolor}
\usepackage[numbers,sort]{natbib} %
\usepackage{wrapfig}
\usepackage{xspace}
\usepackage{stmaryrd}

\usepackage[pagebackref,breaklinks,colorlinks]{hyperref}

\newcommand{\omitme}[1]{}

\newcommand{\method}{Zorro\xspace}

\definecolor{grey}{rgb}{0.5, 0.5, 0.5}
\newcommand\spv[1]{\textcolor{grey}{#1}}

\usepackage{pifont}
\let\originalleft\left
\let\originalright\right
\renewcommand{\left}{\mathopen{}\mathclose\bgroup\originalleft}
\renewcommand{\right}{\aftergroup\egroup\originalright}

\usepackage[capitalize]{cleveref}
\crefname{section}{Sec.}{Secs.}
\Crefname{section}{Section}{Sections}
\Crefname{table}{Table}{Tables}
\crefname{table}{Tab.}{Tabs.}

\def\cvprPaperID{ 000} 
\def\confName{CVPR}
\def\confYear{2023}

\author{
	\renewcommand{\thefootnote}{\fnsymbol{footnote}}
	Adri\`a Recasens\textsuperscript{1}\footnotemark[2]
	\renewcommand*{\thefootnote}{\arabic{footnote}}
	\quad
	Jason Lin\textsuperscript{1}
	\quad
	Joāo Carreira\textsuperscript{1}
	\quad
	Drew Jaegle\textsuperscript{1}
	\quad
	Luyu Wang\textsuperscript{1}
	\quad
	Jean-baptiste Alayrac\textsuperscript{1}
	\\
	Pauline Luc\textsuperscript{1}
	\quad
	Antoine Miech\textsuperscript{1}
	\quad
	Lucas Smaira\textsuperscript{1}
	\quad
	Ross Hemsley\textsuperscript{1}
	\quad
	Andrew Zisserman\textsuperscript{1,2}
	\\
	$^1$\small DeepMind \quad  $^2$VGG, Dept.\  of Engineering Science, University of Oxford \vspace{6pt}
	\\
}

\begin{document}

\title{Zorro: the masked multimodal transformer}

\maketitle

\begin{abstract}
Attention-based models are appealing for multimodal processing because inputs from multiple modalities can be concatenated and fed to a single backbone network – thus requiring very little fusion engineering. The resulting representations are however fully entangled throughout the network, which may not always be desirable: in learning, contrastive audio-visual self-supervised learning requires independent audio and visual features to operate, otherwise learning collapses; in inference, evaluation of audio-visual models should be possible on benchmarks having just audio or just video. In this paper, we introduce Zorro, a technique that uses masks to control how inputs from each modality are routed inside Transformers, keeping some parts of the representation modality-pure. We apply this technique to three popular transformer-based architectures (ViT, Swin and HiP) and show that with contrastive pre-training Zorro achieves state-of-the-art results on most relevant benchmarks for multimodal tasks (AudioSet and  VGGSound). Furthermore, the resulting models are able to perform unimodal inference on both video and audio benchmarks such as Kinetics-400 or ESC-50.
\end{abstract}

\section{Introduction}
	\renewcommand{\thefootnote}{\fnsymbol{footnote}}
	\footnotetext[2]{Correspondence to: Adrià Recasens (arecasens@google.com)}
	\renewcommand*{\thefootnote}{\arabic{footnote}}

\label{sec:intro}
Our perception of the world is inherently multimodal: humans and other animals effortlessly integrate many modalities to build their view of the world \cite{ghazanfar2006neocortex, amedi2017task}. Although multimodal integration can help construct a richer perspective on reality \cite{bavelier2002cross, shams2008benefits}, humans can easily process information and perform tasks even when only a single modality (e.g. sound, vision, or touch) is present \cite{shimojo2001sensory, lacey2014visuo, bola2017task}. 
However, this flexibility is hard to find in perceptual computational models. Architectures for multimodal perception have typically been divided on early fusion, mid-fusion and late-fusion, but most of them need all modalities to be present in order to operate. With human flexibility as an inspiration, in this paper we introduce {\em Zorro}, a multimodal Transformer architecture which is able to operate in both a single-modality and multi-modality setting. This property improves the overall performance of the model while opening the door to off-the-shelf self-supervised pre-training.

Our key architectural innovation in \method is to create separate unimodal and multimodal (fusion) representation streams within a single standard Transformer backbone. We achieve this without engineering the architecture, but instead by applying appropriate masks in all attention operations, resulting in some outputs that only capture  individual modalities and some outputs that capture multimodal information. This has the direct benefit that the model can be applied when a subset of modalities is absent, e.g.\ a model trained on audio and video can be evaluated on audio alone.

\begin{figure*}[t!]
	\centering
	\includegraphics[width=\textwidth]{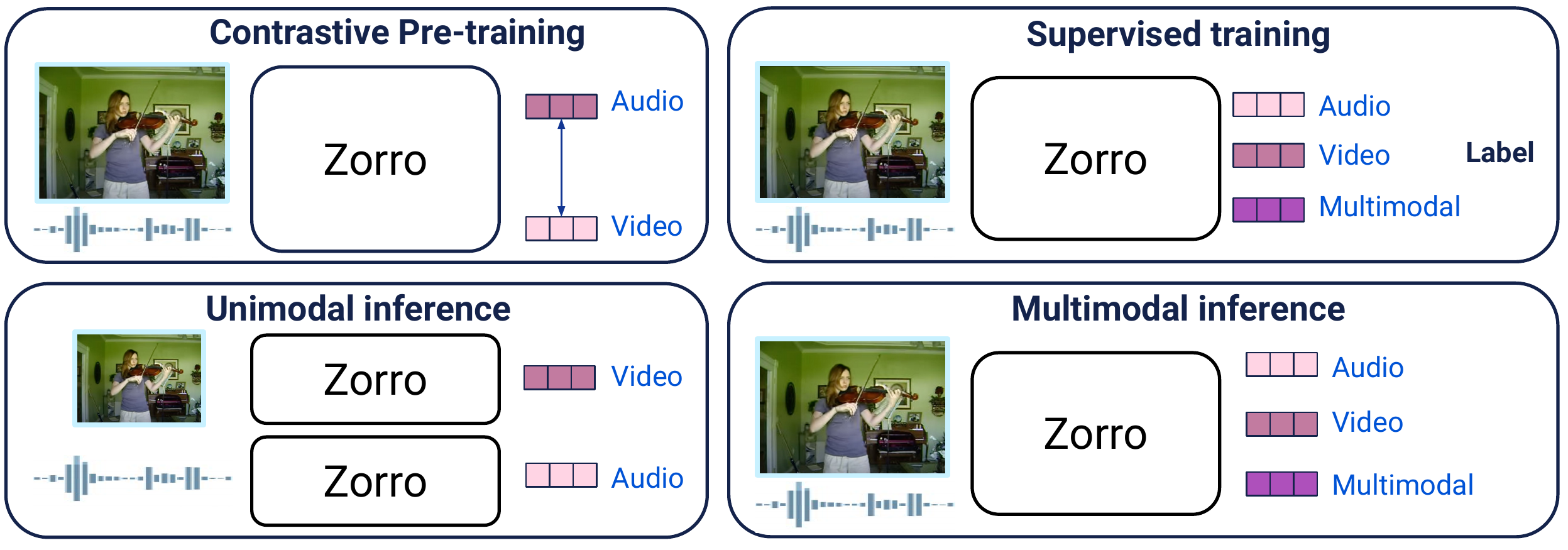}
	\caption{\small 
		In this paper we introduce the \method multimodal architecture which enables both self-supervised contrastive learning and supervised learning. When used for self-supervision, the single-modality outputs are used together with a standard cross-modal self-supervised loss. When used for supervised learning, all outputs can be used for the final classification.
	} 
	\vspace*{-0.3cm}
	\label{fig:teaser}
\end{figure*}

While most of the emphasis of novel developments in the supervised space is put on the architecture, the unimodal outputs can be further exploited by introducing additional self-supervised training schemes. In contrast to recent multimodal attention-based models~\cite{mbt,perceiver} that entangle both modalities throughout the network, \method supports self-supervised contrastive training in a single network without representation collapse, thanks to its unimodal outputs (see Figure~\ref{fig:teaser}). In this work, we explore this possibility by pre-training our model with an audio-visual contrastive loss~\cite{alayrac2020self}. Differently from previous work, we can do this pre-training without the necessity of separate backbones per modality.

This paper presents four contributions: \textbf{(a)} we introduce \method, a novel set of Transformer-based multimodal architectures which enable both supervised and self-supervised training and, once trained, can be used for multimodal or unimodal inputs; \textbf{(b)} we introduce three \method-based architectures using state-of-the-art models such as ViT, SWIN and HiP; \textbf{(c)} we show that \method can be pre-trained on a large-scale audio-visual dataset in a self-supervised manner, and can also be pre-trained on unimodal datasets; and \textbf{(d)} we benchmark our resulting models on AudioSet, VGGSounds, Kinetics-400 and ESC-50. 
The model achieves state-of-the-art performance when compared with previous self-supervised learning techniques on most relevant benchmarks, while also achieving comparable performance with previous work for supervised training with labels.

\section{Related Work}
\label{sec:related}

\noindent \textbf{Multimodal perception}: Multimodal perception is challenging as data from the various modalities can have different topologies, temporal frequencies and relative importances that depend on each task~\cite{baltruvsaitis2018multimodal}. 
With the emergence of convolutional neural networks, numerous works fused activations from intermediate tensors~\cite{wang2020makes,fayek2020large,arandjelovic2018objects,simonyan2014,Feichtenhofer_2016_CVPR,carreira2017quovadis,xiao2020audiovisual}, but this required considerable engineering, as different modalities come in differently shaped feature grids and there are many different ways to combine them. 

\vspace{2mm} \noindent\textbf{Self-supervised audio-visual learning}: Various methods have been used to employ the cross-modality similarity as a self-supervisory signal~\cite{arandjelovic17look,arandjelovic2018objects,Senocak_2018_CVPR,owens2018audio,korbar2018cooperative,alwassel2019self,mandela2020datatrans,morgado20avid}. Most approaches rely on single-modality backbones which produce representations which are used in the self-supervised loss~\cite{alwassel2019self,mandela2020datatrans,alayrac2020self,recasens2021broaden}. These techniques process different modalities with different sets of weights and restrict the ability to reason across modalities. Less common are approaches which learn self-supervised models with multiple modalities at once. One recent work in this direction is \cite{shvetsova2021everything}, which learns representations using audio, video and text. However, to avoid the collapse of the self-supervised loss, they feed the modalities two at a time, increasing the amount of necessary forward passes. Instead, \method masking can produce unimodal outputs without running the model multiple times. 

\vspace{2mm} \noindent\textbf{Transformer architectures}: Inspired by ViT~\cite{vit}, follow up work proposed single-modality processing for video~\cite{vivit} and audio~\cite{gong2021ast} using patch-based encodings. Transformer-based methods have also been proposed to tackle audio-visual classification. The closest to our method is MBT~\cite{mbt}, which builds a multimodal architecture out of single-modality Transformers for video~\cite{vit,vivit} and audio~\cite{gong2021ast}. MBT merges modalities by creating an attention bottleneck which restricts communication between the audio and visual heads. Our method also regulates cross-modality communication, but by masking the latent connections we are able to obtain modality-specific heads while in MBT the representation is entirely multimodal.  Another relevant work is VATT~\cite{VATT}, a Transformer-based architecture to model video, audio and text with a single backbone. Differently from our work, in VATT each modality is independently processed by the transformer. Finally, the Perceiver architecture~\cite{perceiver} scales to a large number of inputs by cross-attending to a set of latent queries. In this work, we use the follow-up Hierarchical Perceiver~\cite{carreira2022hierarchical} which splits inputs and outputs into groups to improve model efficiency. 

\vspace{2mm} \noindent\textbf{Masking attention in Transformers}: The original transformer architecture~\cite{vaswani2017attention} used attention-masking for language modelling. After the success of image-based architectures, alternatives have been proposed to use attention masking to alleviate computational requirements of the architecture. Swin~\cite{liu2021swin} proposed the use of local windows, restricting the self-attention layers to only neighbour pixels. Furthermore, mask2former~\cite{cheng2022masked}, also restricted the cross-attention to local regions, enabling the use of transformers for high dimensional output (e.g segmentation).

\label{sec:method}
	\begin{figure*}[t!]
		\centering
		\includegraphics[width=\textwidth]{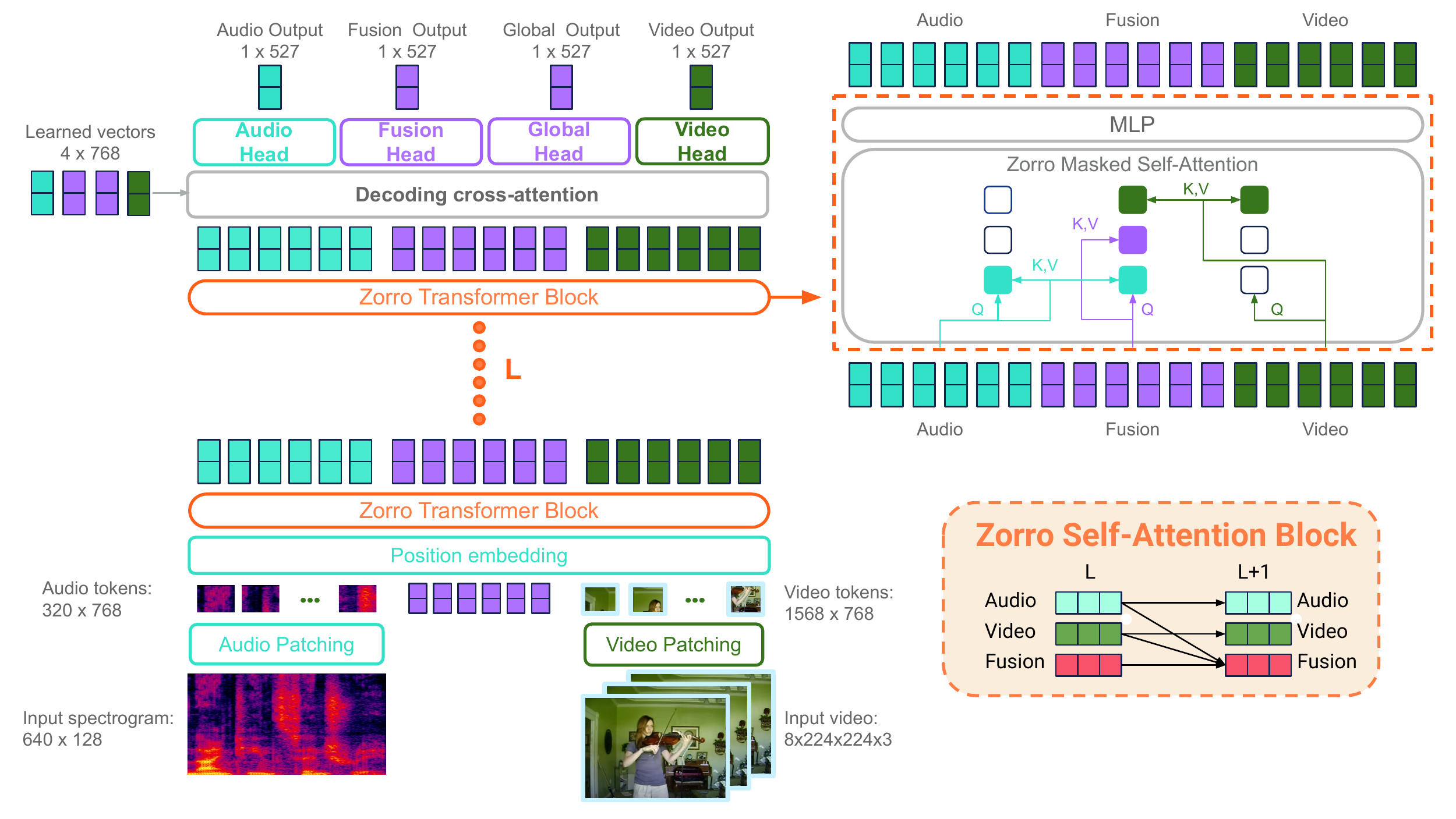}
		\caption{\textbf{The \method-ViT model architecture}: The input to our model are video frames and audio spectrograms. Each of those inputs is patched using a 2D convolution and projection to input dimension $D$. Both audio and video input tokens are concatenated with a set of learned fusion vectors and added a position embedding. Next, we process these inputs through $L$ \method's self-attention layers, where the \method masking is applied. Specifically, our masking strategy blocks the information to flow towards the unimodal hidden representation, while still allowing the fusion representation to access all modalities. By doing this, we ensure that the image and audio representations are gated access to (i.e. depend on) only the video and audio inputs respectively. To produce the outputs, we learn a set of queries that cross-attend (also using masked attention) to the unimodal and multi-modal representation. 
		}
		\label{fig:model_figure}
		\vspace*{-0.05cm}
	\end{figure*}

\section{\method: the masked multimodal Transformer}
\label{sec:model}
In this paper, we introduce \method, a multimodal architecture which enables both supervised and self-supervised training. In this section, we unpack how \method accomplishes this using modality-aware masking and by repurposing the original transformers components to allow contrastive learning between modalities. The key innovation of the architecture is introducing separate latent allocations for the different modalities, leading to a final representation which is partially unimodal (part of the representation sees only a single modality) and partially multimodal (part of the representation can attend to all modalities). First, we will describe \method applied to the ViT architecture. Second, we  extend \method to two other state-of-the-art transformer architectures, Swin and HiP. Finally, we end this section by describing how to use \method for self-supervised contrastive learning. 

\subsection{Architecture}
\label{section:zorro}
\vspace{2mm} \noindent \textbf{\method-ViT overview.}
Figure~\ref{fig:model_figure} depicts the \method architecture, which consist of three main blocks. First, \method processes the data in form of patches (similar to ViT~\cite{vit}). In this stage, data from each modality is first converted into a 2D array of representations. This can be done by either (i) dividing the input tensor into sequential groups (either points or patches) and applying a linear projection, or (ii) applying domain-specific processing such as 1D/2D/3D convolutions and flattening. We use a 2D convolution to extract $16 \times 16$ patches and project them to the input dimension $D$. Next, position embeddings are added to the projected vectors so that the model is able to localise and distinguish each embedded patch.
Learned multimodal fusion vectors are then introduced.
Second, the resulting tokens are concatenated to form a single set and are then processed by $L$ layers of a Transformer~\cite{vaswani2017attention} with \method masking. 
Finally, to produce the final output we learn a set of queries that cross-attend to the output of the last self-attention layer similar to PerceiverIO~\cite{perceiverIO}.
We utilise the standard cross-attention operation~\cite{perceiverIO}, and produce 4 different outputs: an audio output, a video output, a fusion output (which only sees the multi-modal part of the representation) and a global output that sees the whole representation.
These three steps are described in more detail next.

\vspace{2mm} \noindent \textbf{Input pre-processing.} Let $x=(x_v,x_a)$ be a video sample consisting of frames $x_v \in \mathbb{R}^{N_f \times H \times W \times 3}$ and audio spectrogram $x_a \in \mathbb{R}^{T \times N_s}$ where $N_f$ is the number of frames, $N_s$ the dimensionality of the spectrogram, $H$ is the height of the frame, $W$ is the width of the frame and $T$ is the number of temporal steps in the spectogram. To downscale the input, we use a 2D convolution $f^\textrm{patch}$ which yields $u = (u_v,u_a) = (f_v^{\textrm{pre}}(x_v),f_a^{\textrm{pre}}(x_a))$. Arrays $(u_v,u_a)$ are then flattened and absolute learned position encoding are added. Finally, we learn a set of $n_\textbf{fusion}$ latent vectors which are concatenated to the audio and video input tokens. 

\vspace{2mm} \noindent \textbf{Masked attention.} The key contribution of this paper is splitting the Transformer representation into specialised groups. Using masked attention we force part of the representation to attend only to itself, while other parts can attend to the whole representation. The main goal of this approach is to split the representation in three parts: a part which only focuses on video tokens, a part which focuses on audio tokens, and the remaining vectors which can attend to the whole representation. 

We mask two parts of the model: the self-attention~\cite{vaswani2017attention} and the decoding cross-attention~\cite{perceiverIO}. 
Both parts consist of the same underlying operation which takes keys $k$, values $v$ and queries $q$ to produce the final output $o$. To this end, 
we introduce a masking binary tensor $m$ that specifies which vectors are connected to each other.
Entries of the masking matrix are $m_{ij}=1$ if information can flow from latent $i$ to latent $j$. By setting $m_{ij}=0$, we indicate to the model that this connection should be omitted. This mask is applied to the standard attention output operation $o_{i} = \sum_{j} a_{ij} \cdot v_j $ which becomes $o_{i} = \sum_{j} \hat{a}_{ij} \cdot v_j$ where:

\begin{equation}
    \label{eqn:masked}
    \hat{a}_{ij} = \frac{m_{ij} \exp ({\frac{q_i^\top k_j}{\sqrt{D}}})}{\sum\limits_{\{j', \ m_{i{j'}} = 1\}} \exp ( { \frac{q_i^\top k_{j'}}{\sqrt{D}}} ) }.
\end{equation}
In contrast to MBT~\cite{mbt}, our modality-specific representation does not have access to the global representation, which prevents cross-modality information flows. Specifically, we set $m_{ij}=1$ if $j$ is a part of the fusion representation, otherwise we only set $m_{ij}=1$ if $i$ and $j$ are vectors of the same modality. 
By doing this, we explicitly prevent information from the fusion stream leaking into the unimodal representation. 
This is the key to preserving pure streams that correspond to single modalities. 

\vspace{2mm} \noindent \textbf{Output space.}
In ViT architecture, a learnable CLS token is used to produce the output embedding vector. Instead, inspired by the PerceiverIO~\cite{perceiverIO}, we learn a set of decoding vectors which are used to query the output from the Transformer to produce the final output. Each decoding vector cross attends to a subset of tokens to produce the final output vector. This decoding strategy can be used to produce as many outputs as desired, opening up the possibility for dense tasks such as segmentation or flow estimation. 

As we are relying on having the Transformer representation split into specialised groups, we need to also apply \method's masking to the output cross attention. Specifically, we found it beneficial to define four outputs for our model. The audio-specific output $o_A$, which only contains information coming from the audio input. The video-specific output $o_v$, which only includes information from the video modality. The fusion specific output $o_F$, which is computed by attending only to the fusion stream. And finally, a global output $o_G$, which attends to all the outputs in the model. Although $o_G$ and $o_F$ do contain similar information, we found it useful to still keep two different heads.

\subsection{Extending \method for other architectures}
In this section, we propose variants of \method for two state-of-the-art attention-based architectures, Swin and HiP. Differently from the ViT implementation, when building  \method-Swin and \method-HiP we use the specific architecture building block for each modality and the fusion stream while we join the modalities with a cross-attention operation. This is required as the ViT masking is not directly applicable to Swin and HiP, but the overall idea remains the same. 

\noindent \textbf{\method-Swin}: 
Swin~\cite{liu2021swin} is a ViT-inspired transformer architecture which has shown improved efficiency and performance. The main innovation versus the original ViT architecture is to apply the self-attention operations on nearby tokens instead of all tokens in the input image. This reduces computational requirement while allowing the model to perform bottom-up inference. In order to build \method-Swin, our main modification to the original architecture is to process individual modalities using Swin transformers. At the end of each Swin block, we update the fusion representation by cross-attending to both the unimodal and multimodal representation. To process the fusion representation, we use the same self-attention as in \method-ViT. Given this design, we are free to use different architectures to process each modality. We use the original 2D Swin~\cite{liu2021swin} to process the audio spectrograms while our adaptation of the Swin architecture for video. Similarly to \method-ViT, no multimodal information flows into the unimodal streams. Detailed description of \method-Swin can be found in Section~\ref{arch:details} in the Appendix.

\noindent \textbf{\method-HiP}: The hierarchical perceiver~\cite{carreira2022hierarchical} extends the previously introduced Perceiver models~\cite{perceiver, perceiverIO} models, by splitting the inputs into groups, and operating only within those groups. Through the hierarchical architecture, those groups fuse together in order to aggregate information and globally reason about the input. In our implementation of HiP, instead of using directly the pixels and audio signal as input, we create patches similarly to the ViT/Swin implementation. In order to create \method-HiP, we use HiP building blocks for each modality. Specifically, those blocks group the inputs into smaller sets, cross-attend using learned features and finally apply self-attention layers to the outputs of the cross attention operation (see~\cite{carreira2022hierarchical} for more details). In order to update the fusion representation, we learn a set of queries which cross attend to both unimodal and multimodal representation per each layer. More details can be found in Section~\ref{arch:details} in the Appendix.

\subsection{Contrastive learning with \method}
\label{sec:selfsup}

Contrastive audio-visual methods learn representations by aligning audio and video into a common embedding space. As opposed to unimodal approaches, instead of producing multiple views of the data, they use different modalities as views. 
One important requirement is for the two backbones to not share information. If information is shared across modalities, the self-supervised training can easily collapse or converge to a trivial solution.

Models for multimodal perception typically produce a single output for the multiple inputs. This is sufficient for supervised applications, but prevents the use of these audio-visual contrastive techniques. We design \method in order to process unimodal and multimodal outputs, with the intention of enabling the use of self-supervised contrastive losses. 

\vspace{2mm} \noindent \textbf{Noise Contrastive Estimation}: For training with the standard noise-contrastive estimation loss, we follow the implementation of the audio-visual loss 
from~\cite{alayrac2020self}. Given the audio output $o_a$ and the video output $o_v$, we apply a final linear projection (different per modality) $g_a$ and $g_v$ to yield the final embedding vectors: $z_a = g_a(o_a)$ and $z_v = g_v(o_v)$. We compute the similarity between $z_a$ and $z_v$ by taking a normalised dot product and dividing by a temperature parameter $\tau$, $\textrm{sim}(z_a,z_v) = \exp (\frac{\hat{z_a} \hat{z_v}}{\tau})$. Finally we apply the NCE loss:
\begin{equation}
    L_{\textrm{NCE}}(z_a,z_v) =- \sum_i \log \frac{\textrm{sim}(z_a^i,z_v^i)}{\sum_{j,k} \textrm{sim}(z_a^k,z_v^j)}
    \label{eqn:nce}
\end{equation}

Equation~\ref{eqn:nce} introduces describes the loss for audio-visual contrastive training. However, this technique does not train any parameters specific to the fusion representation or output (e.g, the fusion cross-attention or the fusion weights if the model has separate weights per modality). In order to self-supervise the output of the fusion stream, we add a fusion-visual and fusion-audio contrastive loss. We define a self-supervised loss contrasting both unimodal representations (audio and video) separately with the multimodal one (fusion). With those changes, the new loss is:
\begin{equation}
    \label{eqn:fusion_contrastive}
    \small L_{\textrm{NCE}} = L_{\textrm{NCE}}(z_a,z_v)+L_{\textrm{NCE}}(z_a,z_f)+L_{\textrm{NCE}}(z_I,z_f)
\end{equation}

\section{Experiments}
\label{sec:experiments}
In this section, we evaluate the \method architecture on multiple settings. We first present details of the training and evaluation procedures, as well as the main datasets we use. We evaluate the method against state-of-the-art models on three standard audiovisual benchmarks ( AudioSet~\cite{gemmeke2017audio}, VGGSound~\cite{chen2020vggsound} and Kinetics-400~\cite{carreira2017quovadis}), one vision benchmarks (Kinetics-400~\cite{carreira2017quovadis}) and one audio benchmark (ESC-50~\cite{piczak2015dataset}). Finally, we ablate the main design decisions that drove our research and showcase \method's flexibility. Specifically, we compare the different architectures, study the effect of missing modalities, pre-train \method with unimodal data and explore alternative attention-masking strategies.

\subsection{Experimental details}
In order to showcase \method's ability to reason across different modalities, we pre-train it using self-supervision as well as with standard supervision using class labels. 
In this section, we provide the most important details of the training procedure. Additional details about inputs, architectures and training can be found in Section \ref{arch:details} and \ref{sec:training_details} in the Appendix.

\vspace{2mm} \noindent \textbf{Pre-training datasets}:  We utilise four datasets for pre-training: AudioSet~\cite{gemmeke2017audio}, YouTube-8M, ACAV-100M~\cite{lee2021acav100m} and ImageNet-21k~\cite{ridnik2021imagenet}. AudioSet consist of $1.9$M videos which contain $527$ classes of annotated sounds. As the dataset is highly unbalanced, \cite{mbt} proposed a smaller more balanced variant of the training set with $500$k examples. For the ablation experiments and training from scratch, we use the $1.9$M version while for fine-tuning we also use AudioSet-500k for fair comparison with the state-of-the-art. YouTube-8M~\cite{abu2016youtube} consist of $8$M videos with audio and visual frames, annotated in a multi-label fashion with $3862$ different classes. Videos are representative of many activities, resulting a very natural distribution of data. ACAV-100M consist of $100$M videos with audio and visual frames without associated labels, which have been curated to contain a strong audio-visual correlation. We use $59M$ of those videos for self-supervised learning.  ImageNet-21k consist of $13M$ images annotated on $21k$ classes, and been typically used for large-scale pretraining of visual transformer models~\cite{vit}.

\vspace{2mm} \noindent  \textbf{Audio-visual evaluation benchmarks}: To evaluate the ability of \method to learn and transfer multimodal representations, we evaluate on standard audio-visual benchmarks. Specifically, we evaluate \method in AudioSet, VGGSound~\cite{chen2020vggsound} and Kinetics-400~\cite{kay2017kinetics}. VGGSound consists of $163,603$ training and $13579$ test samples drawn from 10-second YouTube videos which span $309$ single-label, mutually exclusive classes. It focuses on real life audio evaluation with audio-visual correspondence where sounds are visually evident in the video. 
Kinetics-400 consists of $201$K training videos of everyday actions which are classified into $400$ unique classes. While some datasets have bias in audio or video modality, \method is able to learn the extent to rely on each modality. 

\vspace{2mm} \noindent  \textbf{Unimodal evaluation benchmarks}: \method can be trained on multi-modal data but evaluated on unimodal data. To further show this we evaluate the multi-modal trained \method models on unimodal fine-tunning tasks: Kinetics-400 for vision and ESC-50 for audio. ESC-50 dataset contains $2k$ clips classified into $50$ unique classes.

\vspace{2mm} \noindent\textbf{\method inputs}: The inputs to our model are video and audio. The audio and video are synced and cover the same time span. Video consists of $8$ frames of size $224 \times 224$. When training in AudioSet, we sample videos at $3.12$FPS which results on $2.56s$ of audio and video. Specific FPS per model and audio length for pre-training and fine-tuning is reported in Section~\ref{sec:training_details} in the Appendix.
During training, we use random cropping as well as color augmentation in frames. 
For ESC-50, we match the lengths of the pre-trained model, looping over the audio sequence if required. Audio is sampled at $48kHz$, converted to spectrograms as inputs to our model using $128$ bins.
To augment the audio in training, we use SpecAugment~\cite{park19specaug} and frequency jittering. During evaluation, we subsample the input video and audio into multiple equally spaced clips and averge their predictions.

\vspace{2mm} \noindent\textbf{Architectural details}: 
\method is based on unimodal transformer architectures (ViT, Swin and HiP), adapted for multimodal processing (similar to~\cite{mbt}). Through all our experiments we use ViT-B/16. For details on ViT, Swin and HiP architecture, see Section~\ref{arch:details} in the Appendix.

\vspace{2mm} \noindent\textbf{Training details}:
We use the Adam optimiser with cosine decay learning rate schedule, weight decay and learning rate warmup. When fine-tuning, for \method-ViT and \method-Swin we find better to use SGD optimiser and momentum $0.9$. We train all models for $50$ epochs except for the ACAV-100M datasets where we train for $10$ epochs and the \textit{input-level} and \textit{bottleneck} baselines where we train for $25$ to prevent severe overfitting. We find best to use $n_\textbf{fusion}=6$ in all models. For  AudioSet fine-tuning, we use mixup ($\alpha=0.3$) and label smoothing. We use cross-entropy loss for uni-label datasets and binary sigmoid cross-entropy for multi-label. We train one classifier for each of the $4$ outputs of the model and average its predictions. For contrastive training, we follow the procedure outlined in Section~\ref{sec:selfsup}.

\subsection{State-of-the-art comparison}
Next, we evaluate \method against state-of-the-art methods. We evaluate our audio-visual trained \method on benchmarks for audio-visual classification, video classification and audio classification, showcasing the universality of the approach.

	\begin{table}[t]
		\centering
		\caption{\small {\bf AudioSet-2M comparison: training from scratch.} We report the performance of our models trained on audio-visual data compared with the state-of-the-art when trained from scratch. We report the mean average precision on the AudioSet test set.
        }
			\begin{tabular}{c|cc|c} \toprule
			Model  & Train Mod & Eval Mod & AudioSet \\
			 \hline
			 HiP~\cite{carreira2022hierarchical} &  A+V &  A+V & 43.8\\
			 Perceiver~\cite{perceiver} &  A+V &  A+V & 44.2\\
			 ERANN~\cite{verbitskiy2021eranns} & A &  A & 45.0 \\
            \hline
			 \method-ViT &A+V &  A+V & 45.1 \\ 
			 \method-HiP & A+V &  A+V & 45.2 \\ 
			 \textbf{\method-Swin} & \textbf{A+V} &  \textbf{A+V} & \textbf{46.5} \\

			\bottomrule
		\end{tabular}
		\label{tab:audioset_scratch}
	\end{table}

	\begin{table*}[t]
		\centering
		\caption{\small {\bf State-of-the-art results:} We compare \method with the state-of-the-art in two settings: when labels are not used in pre-training or when labels are used. We report the mean average precision on the AudioSet test set and top-1 accuracy on K-400, VGGSound and ESC-50. IN-21k is ImageNet-21k~\cite{ridnik2021imagenet}, YT8M is YouTube-8M~\cite{abu2016youtube}, ACAV is ACAV-100M~\cite{lee2021acav100m} and K-400 is Kinetics-400~\cite{carreira2017quovadis}. When using ImageNet-21k initialisation, we use the pre-trained weights to initialise the video, audio and fusion parameters.}
			\begin{tabular}{c|ccc|ccc|c|c} \toprule
  			Model & \multicolumn{3}{c|}{Pre-Training} & \multicolumn{3}{c|}{Eval: Video+Audio}   & Eval: Video & Eval: Audio  \\      
            \hline
			 & Dataset & Sup/SSL   & Mod & AS & VGGSound & K-400 & K-400 & ESC-50 \\
			 \hline

  			 \multicolumn{4}{l|}{\textbf{No pre-training}} & &    &  & &   \\         

			 SlowFast R101-NL~\cite{feichtenhofer2019slowfast} & &  &  & & & \bf 79.8 & \bf 79.8 \\
			 AVSlowFast~\cite{xiao2020audiovisual}, R101 & & & & & & 78.8  &  \\

			 AudioSlowFast~\cite{kazakos2021slow} &  &  & & & 52.5 &  &  \\
			 ERANN~\cite{verbitskiy2021eranns}, R101 & & & & 45.0& &   & &  89.2 \\

			 PlayItBack~\cite{stergiou2022play}, R101 & & & & 47.7& 53.7&   &  \\
             \midrule

  			 \multicolumn{4}{l|}{\textbf{Self-supervised pre-training}} & &    &  & &   \\         

			 MaskSpec~\cite{chong2022masked}, ViT & AS &SSL & A& 47.1& & &   &89.6  \\

			 \method-HiP  &ACAV & SSL&  A+V  &49.4 & 61.3 &67.9 &  64.6 & 88.4\\
			 \method-Swin  &ACAV & SSL  &  A+V  &49.4 & 61.1& 73.7 &  69.4 & 91.4 \\

			 \method-ViT  &ACAV & SSL  &  A+V  & \bf 50.3  & \bf  63.6 & 76.5 &74.1 & \bf 93.6\\

             \midrule
  			 \multicolumn{4}{l|}{\textbf{Supervised pre-training}} & &    &  & &   \\         

			 \spv{ViViT-Base~\cite{vivit}} & \spv{IN-21k} & \spv{Sup.} & \spv{V} &  & &\spv{80.0} &  \spv{80.0} & \\
			 \spv{MaskSpec~\cite{chong2022masked}, ViT} & \spv{AS} &\spv{Sup} & \spv{A}& & &   && \spv{98.2}  \\

			 \spv{ PaSST}~\cite{koutini2021efficient} & \spv{IN} &\spv{Sup.} &  \spv{V}   & \spv{49.6}  & &  & &\spv{96.8} \\
			 \spv{ AST~\cite{gong2021ast}} & \spv{IN-21k} & \spv{Sup.} &  \spv{V}   & \spv{45.9} & & &  &  \spv{95.7}\\

			  \spv{MBT~\cite{mbt},ViT}  & \spv{IN-21k} & \spv{Sup.}  & \spv{V} & \spv{49.6} & \spv{64.1} & \spv{80.8} &\spv{79.4}  \\
			 \spv{\method-ViT} & \spv{IN-21k} & \spv{Sup.} & \spv{V} &  \spv{50.9} & \spv{63.1} & \spv{79.8}  &\spv{77.6}  & \spv{81.7}\\

			  \spv{\method-ViT} & \spv{YT8M} & \spv{Sup.} & \spv{A+V} &  \spv{51.5} & \ \spv{64.8} &\spv{79.6} & \spv{76.1}&\spv{93.1}   \\

			\bottomrule
		\end{tabular}
		\label{tab:sota_experiments}
	\end{table*}

\vspace{2mm} \noindent \textbf{Training AudioSet-2M from scratch}: First, we evaluate \method when trained from scratch on Audioset-2M using both the audio and visual modalities. Table~\ref{tab:audioset_scratch} reports that \method matches or overperforms other methods that directly trained on AudioSet-2M from scratch. Note that PlayItBack~\cite{stergiou2022play} is not listed in Table~\ref{tab:audioset_scratch} as it was trained with AudioSet-500k. This setting shows the model's ability to adapt to the multi-modal inputs without the need of pre-trained data. 

\vspace{2mm} \noindent \textbf{Multi-modal comparison}: We train and evaluate our pre-trained models on AudioSet-500k (see \cite{mbt} for details), VGGSound and Kinetics-400 where we use both the audio and visual inputs. Similar to \cite{mbt}, for \method-ViT we allocate different weights for the audio, video and fusion latents. We found this useful for improving the fine-tuning accuracy. 
Table~\ref{tab:sota_experiments} reports the performance of our models. 
We divide the table into two different parts. First, we report the \method performance when contrastive self-supervision is used for pre-training (no labels). \method improves over all previous works on AudioSet and VGGSound. In AudioSet, our best-performing model on that setting is only $1.2\%$ away from \method with supervised pre-training, which demonstrates the ability of the self-supervised pre-training technique for learning general features. In VGGSound, \method performs similarly with the supervised state-of-the-art when pre-trained only with self-supervision. Finally, for Kinetics-400, the resulting performance is not far from models with supervised pre-training. 
In the bottom part of the table we report the performance of \method when using supervised pre-training. We include the performance of the model when initialized with ViT pre-trained on ImageNet-21k. Even without multi-modal pre-training, \method is able to perform more than $1\%$ better than existing SOTA models in AudioSet. When pre-trained on YouTube-8M, \method improves its performance as a result of the multi-modal nature of its pre-training. The final performance in AudioSet represents a $1.9\%$ improvement over the state-of-the-art, MBT~\cite{mbt}. Furthermore, unlike \method, MBT cannot perform unimodal inference when trained with multi-modal data. 
Note, we have not demonstrated it here, but \method can also be trained using unimodal self-supervised methods such as MAE~\cite{he2022masked} and DINO~\cite{caron2021emerging} separately on the audio and visual streams. We discuss supervised unimodal training below.

\vspace{2mm} \noindent \textbf{Video comparison}: To showcase \method's performance in the unimodal regime, we fine-tune our models (pre-trained on audio and video) on the task of video classification for Kinetics-400 using only video. Table~\ref{tab:sota_experiments} reports the results. Our goal is not to show state-of-the-art performance on this setting, as we are aware of the improvements made on Transformer architectures to solve that task~\cite{zhang2021co,liu2021swin,yan2022multiview}. Our goal is to provide an efficient mechanism for pre-training those architectures in order to improve the final performance on unimodal and multimodal inference. When \method is pretrained using a contrastive loss and fine-tuned on Kinetics-400 (video only), \method-ViT performs only $2.4\%$ worse than when using audio-visual input. This shows the robustness of our model when reduced to using a single modality. Furthermore, when using the \method model pre-trained on YT8M, our model is able to perform similarly to comparable architectures. Alternative to fine-tuning, we can also use the audio-visual trained model (column \textit{Audio+Video}) and only feed the video. In that setting, our model trained on YouTube-8M performs at $76.3$ top-1, on par with the video only fine-tuned result. This unimodal inference on a multi-modal trained model is not possible with MBT, where retraining is needed. 

\vspace{2mm} \noindent \textbf{Audio comparison}: To evaluate \method's audio capabilities, we fine-tune our models on ESC-50 (audio-only dataset) and report results in Table~\ref{tab:sota_experiments}. When pre-trained on YouTube-8M, \method performs close to AST, an specialised audio transformer comparable in size. When using self-supervised pre-training, \method improves performance over previous methods; \method-ViT has an accuracy of $93.6\%$, close to state-of-the-art supervised methods. 

	\begin{table*}[t]
		\centering
		\caption{\small {\bf Masking configurations and architectures:} We evaluate the different masking configurations by training \method on AudioSet with a supervised loss and audio-visual contrastive loss. Specifically, we test the audio-visual trained models on a unimodal (Audio, Video) and multimodal setting. Our proposed configuration performs well across the board while providing additional unimodal outputs.
        }
			\begin{tabular}{ccc|ccc|ccc} \toprule
			& & &\multicolumn{3}{c}{Supervised (Audio+Video)}   & \multicolumn{3}{c}{Self-Supervised (Audio+Video)}  \\
			\hline

			Architecture & Params &  Fusion & Video & Audio   & Audio+Video & Video & Audio & Audio+Video \\
			\hline

			ViT & 98M & Two Streams  &23.1 & 40.1&  42.2  &18.9 &32.3 &34.8  \\
			ViT & 98M &  Input Level  &9.1 & 31.6&  42.2  &Collapse & Collapse&Collapse  \\
			ViT & 98M & Bottleneck ~\cite{mbt}  &9.7 & 32.6&  42.5  &Collapse & Collapse&Collapse  \\
			ViT & 98M &  \method & 22.5 & 39.7&  45.1  &17.8 &29.8 &33.6  \\
			HiP & 136M &   \method &22.0  & 39.5& 45.2   &11.3 &21.9 & 26.5  \\
			Swin & 161M &   \method & 25.4& 40.6& 46.5  &20.5 & 31.6& 35.7 \\

			\bottomrule
		\end{tabular}
		\label{tab:ablation_experiments}
	\end{table*}
\subsection{Architecture comparison}
In this section, we discuss the different architectures introduced in this paper. 
In Table~\ref{tab:ablation_experiments} we report comparison for those architectures in two settings: when trained from scratch and when pre-trained with an audio-visual contrastive loss followed by a linear layer on top, using Audioset-2M. 
When training from scratch, we observe \method-Swin performs the best across the different models, both in the supervised and contrastive regimes. Although the number of parameters is larger than ViT, Swin trains $25\%$ faster than ViT. HiP is the fastest of the three, while not losing much on accuracy. See Section~\ref{arch:details} in the Appendix for model speed comparison. Furthermore, in Table~\ref{tab:sota_experiments} we also present the results of fine-tuning these architectures after contrastive pre-training. It is important to note that for ViT, in this table we use one set of parameters per modality, which significantly increases the parameter count ($98$M to $267$M). In this regime, we observe how ViT is the best. However, Swin and HiP are faster and retain most of the performance. 

\subsection{Zorro model flexibility}

\vspace{2mm} \noindent \textbf{Unimodal inference with a multimodal backbone}:
Here we study the ability of audio-visual trained \method to produce meaningful unimodal outputs when fed with unimodal data.
To achieve this we zero out the missing modality and only provide useful inputs for one modality, either video or audio. 
Results are reported in Table~\ref{tab:ablation_experiments}. Models without unimodal output suffer significantly from one missing modality. In contrast, both \method and using two separate modality streams achieve a high performance when only a single modality is provided. This is due to the fact that in those models, some capacity is allocated to each modality specifically and the model is able to produce unimodal outputs.

\vspace{2mm} \noindent \textbf{Unimodal pre-training for multi-modal fine-tuning}:
Through the paper, we assumed availability of large multi-modal dataset for training. However, in some situations we only have available large amounts of unimodal samples (e.g.\ video or audio) and a small set of multi-modal data. To showcase the flexibility of our proposal, we run a single experiment where we train with two unimodal datasets and fine-tune on a smaller multi-modal dataset. We use only the audio signal from the AudioSet dataset and the videos from the Kinetics-400 dataset. When training, we mix batches with probability $0.5$ per dataset, and do not compute the loss for the missing modalities. For evaluation, we fine-tune the resulting model on VGGSound and compare its result to the model trained from scratch. The fine-tuned model performs at $59.2$ top-1 accuracy while the model trained from scratch performs at $54.4$. This experiment shows the flexibility of the \method model to adapt to unimodal training while providing useful initialization for multi-modal fine-tuning.

\subsection{Masking configurations}
In this ablation, we study four different types of attention masking. First, we evaluate having data independent stream (\textit{two streams}), where both models share weights but modalities are not connected. Secondly, we evaluate input level fusion, which consist of no masking in the model. This reduces the model to a vanilla ViT applied to the two concatenated modalities. Inspired by~\cite{mbt}, we also evaluate \textit{bottleneck masking} where the fusion tokens can attend to each modalities' tokens but each modality can also attend to the fusion tokens. We want to make clear that although this approach uses the main proposal from MBT, it is not a reproduction of their work. This configuration forces each stream to mostly concentrate on one modality, but information can flow across modalities through the fusion vectors. Finally, we compare all those masking strategies with our \method masking. For each masking configuration we train a model in a supervised manner (keeping the same number of outputs for fairness, except for the Two Streams which has two outputs). We also train the model in a self-supervised way, where the audio and the video outputs are used to compute the contrastive loss. To report performance, we train a linear classifier on top of the contrastive representations.

Table~\ref{tab:ablation_experiments} reports the results. We extract two main conclusions. First, having modality independent streams is crucial for self-supervised training. Both the \textit{input-level} and the \textit{bottleneck} configurations immediately collapse as information can flow from one modality to the other. Performance for \method and \textit{two streams} is very similar as \method when trained in a self-supervised manner reduces to the two stream architecture. Secondly, we find that having separate modality streams is useful also for supervised learning. Specially interesting is looking at the performances of \textit{input-level}, \textit{bottleneck} and \method, where \method performs better as the modality streams are more independently treated. We believe this is due to the ability of the model to keep modality-specific information through the network, which can be useful at later stages of processing. Finally, for self-supervised training of \method, we use equation~\ref{eqn:fusion_contrastive}, which trains also the fusion output. Although this produces a slight decrease on performance vs \textit{two streams}, it's beneficial for downstream tasks. Alternatively, when \method is trained using only audio and video outputs, it performs the same as \textit{two streams} ($35.0$ vs $34.8$) as the two models are equivalent.

\section{Conclusion}
\label{sec:conclusions}
In this paper, we introduced \method, a novel Transformer masking configuration which enables simultaneous unimodal and multimodal training and inference, as well as contrastive pre-training. Different from previous approaches to multimodal perception, our proposed method is able to generate both unimodal and multimodal outputs. By splitting the information flow into unimodal and multimodal streams, we are able to improve performance when the architecture is trained with a supervised loss and show the ability of the model to be self-supervised with a contrastive loss. We evaluate our model on multimodal tasks, showing great flexibility and state-of-the-art performance.

\bibliographystyle{ieee_fullname}
\bibliography{egbib}
\clearpage
\pagebreak

\appendix

\section*{Appendix}
In this appendix we expand the content of the paper in three different directions. First, we describe in detail the three Zorro architectures proposed in the paper and compare their speed. Second, we provide details about training, evaluation and input pre-processing. Third, we study the importance of the level at which the modalities are fused.

\section{Architecture details}
\label{arch:details}
\vspace{2mm} \noindent\textbf{ViT}: \method -ViT is based on the ViT-B/16 architecture, adapted for multi-modal processing (similar to~\cite{mbt}). For input video and audio frames, we use patch size $16 \times 16$ and hidden dimension of $768$. The model has $12$ self-attention layers, with intermediate dimension $3072$ and $12$ attention heads with $n=6$ fusion tokens. For decoding cross-attention, we use a decoder with $q_{k}$ dimension $1024$. Figure~\ref{fig:model_figure}  details the \method-ViT architecture. We use absolute learned position embedding.

\noindent \textbf{\method-Swin}: 
Swin~\cite{liu2021swin} is a ViT-inspired transformer architecture which has shown improved efficiency and performance. The main innovation versus the original ViT architecture is to apply the self-attention operations on nearby tokens instead of all tokens in the input image. This reduces computational requirement while allowing the model to perform bottom-up inference. In order to build \method-Swin, we adapt the original 2D Swin~\cite{liu2021swin} to deal with video data by adding a third dimension to the attention window (we use $3 \times 7 \times 7$) and the position encoding. Similarly to the 2D version, input tokens are only attended locally, and windows shift by half window size every two layers.

\begin{figure*}[t!]
		\centering
		\includegraphics[width=\textwidth]{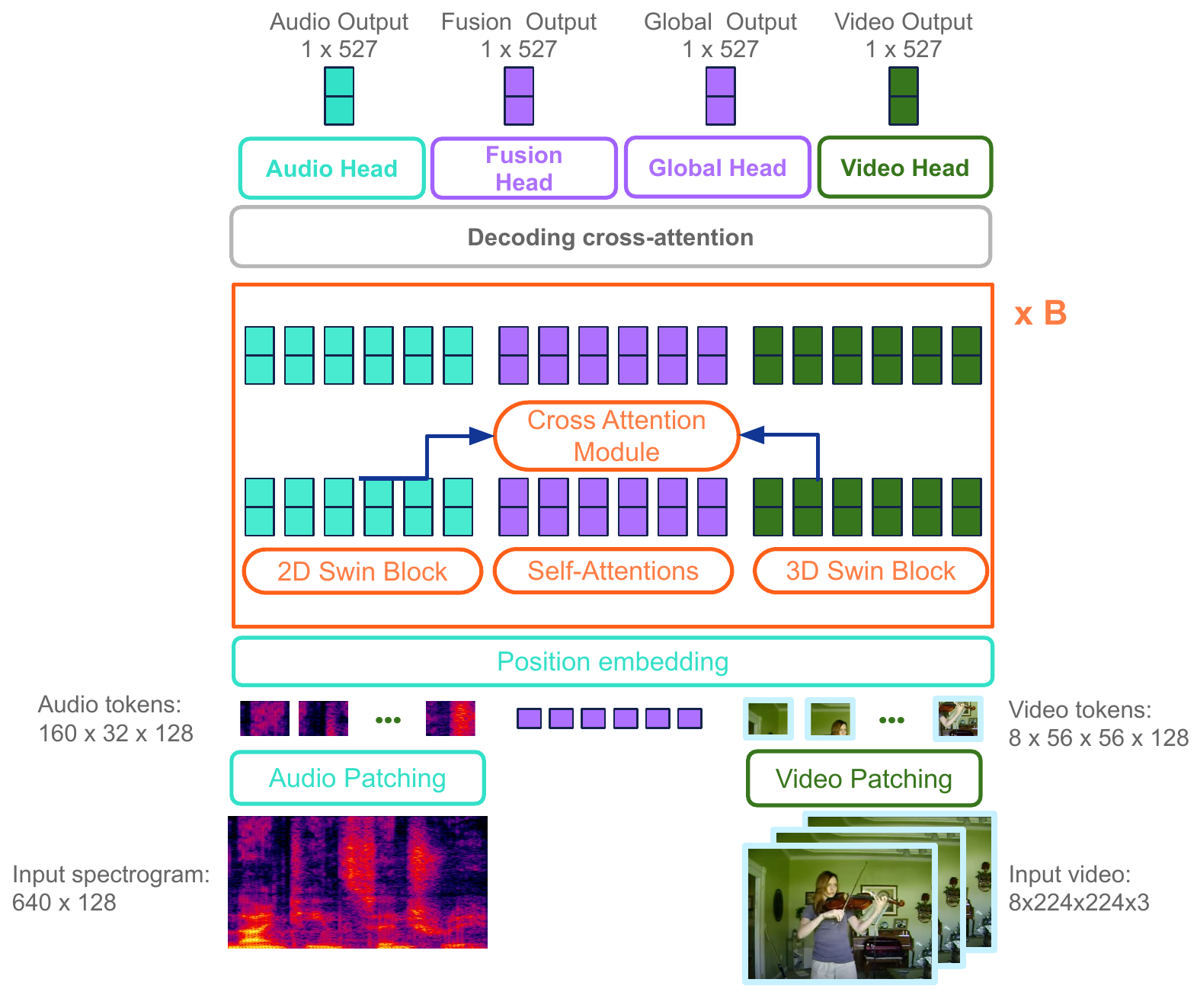}
		\caption{\textbf{\method-Swin}: The input to our model are video frames and audio spectrograms. Each of those inputs is patched using a 2D convolution and projection to input dimension $D=128$. Next, the audio tokens are processed by a 2D Swin, while the video tokens are processed by a 3D Swin. The fusion tokens are processed by standard self-attention layers. At the end of each of the $B$ blocks, a cross-attention operation is applied to produce the next fusion tokens. Specifically, our architecture blocks the information to flow towards the unimodal hidden representation, while still allowing the fusion representation to access all modalities. By doing this, we ensure that the video and audio representations have gated access to (i.e.\ depend on) only the video and audio inputs respectively. To produce the outputs, following Perceiver IO~\cite{perceiverIO}, we learn a set of queries that cross-attend to the unimodal and multimodal representation. We also use masking at the decoding stage to make sure we can produce unimodal outputs as well as multimodal outputs. By doing this, we can train \method-Swin using a self-supervised loss which requires unimodal representations. 
		}
		\label{fig:swin_figure}
		\vspace*{-0.05cm}
	\end{figure*}

The \method-Swin architecture is depicted in Figure~\ref{fig:swin_figure}. As with \method-ViT, the model processes independently audio and video, while the fusion tokens cross-attend to the whole representation. The main difference is that to process video and audio, we use a Swin model. Specifically, to process the input audio (spectrograms) we use the original 2D Swin~\cite{liu2021swin}, while for video we use our 3D Swin architecture adaptation. For 2D Swin we use $(4,4)$ patches, while for 3D Swin we use $(1,4,4)$. Furthermore, we use $6$ fusion tokens (as in \method-ViT) and the input embedding dimension is $128$. The number of layers per block is $(2,2,6,2)$ in both cases, number of heads are $(4, 8, 16, 32)$, MLP dimensional ratio is $4$ and we use stochastic layer drop with probability starting at $0$ in the first layer and linearly increasing to $0.3$ in the last layer. To process the fusion representation, we use the same self-attention as in \method-ViT with $16$ heads and widening factor of $4$. We use the relative position bias from \cite{liu2021swin}. Similarly to \method-ViT, no multimodal information flows into the unimodal streams. 

\noindent \textbf{\method-HiP}: The hierarchical perceiver~\cite{carreira2022hierarchical} extends the previously introduced Perceiver models~\cite{perceiver, perceiverIO} models, by splitting the inputs into groups, and operating self-attention only within those groups. Through the hierarchical architecture, those groups fuse together in order to aggregate information and globally reason about the input. 

In our implementation of HiP, instead of using directly the pixels and audio signal as input, we create patches similarly to the ViT/Swin implementation. Specifically, inspired by~\cite{touvron2022three}, we produce the input patches by processing the input through a sequence of two (Convolution + LayerNorm + GeLU) operations and a final (Convolution + Layer Norm) at the end. The initial two convolutions project the input to $64$ dimensions and the last one to $256$ dimensions. The initial convolution has stride $(2,2,2)$ when processing video and $(1,2,2)$ when processing audio. The other convolutions have stride $(1,2,2)$. The final downsample is $(2,8,8)$ for video and $(1,8,8)$ for audio. We add Fourier positional features and afterwards a learned embedding. Although the model does not need both positional features, we add the learned positional embedding to make the model as similar as possible to other \method variants.

	\begin{figure*}[t!]
		\centering
		\includegraphics[width=\textwidth]{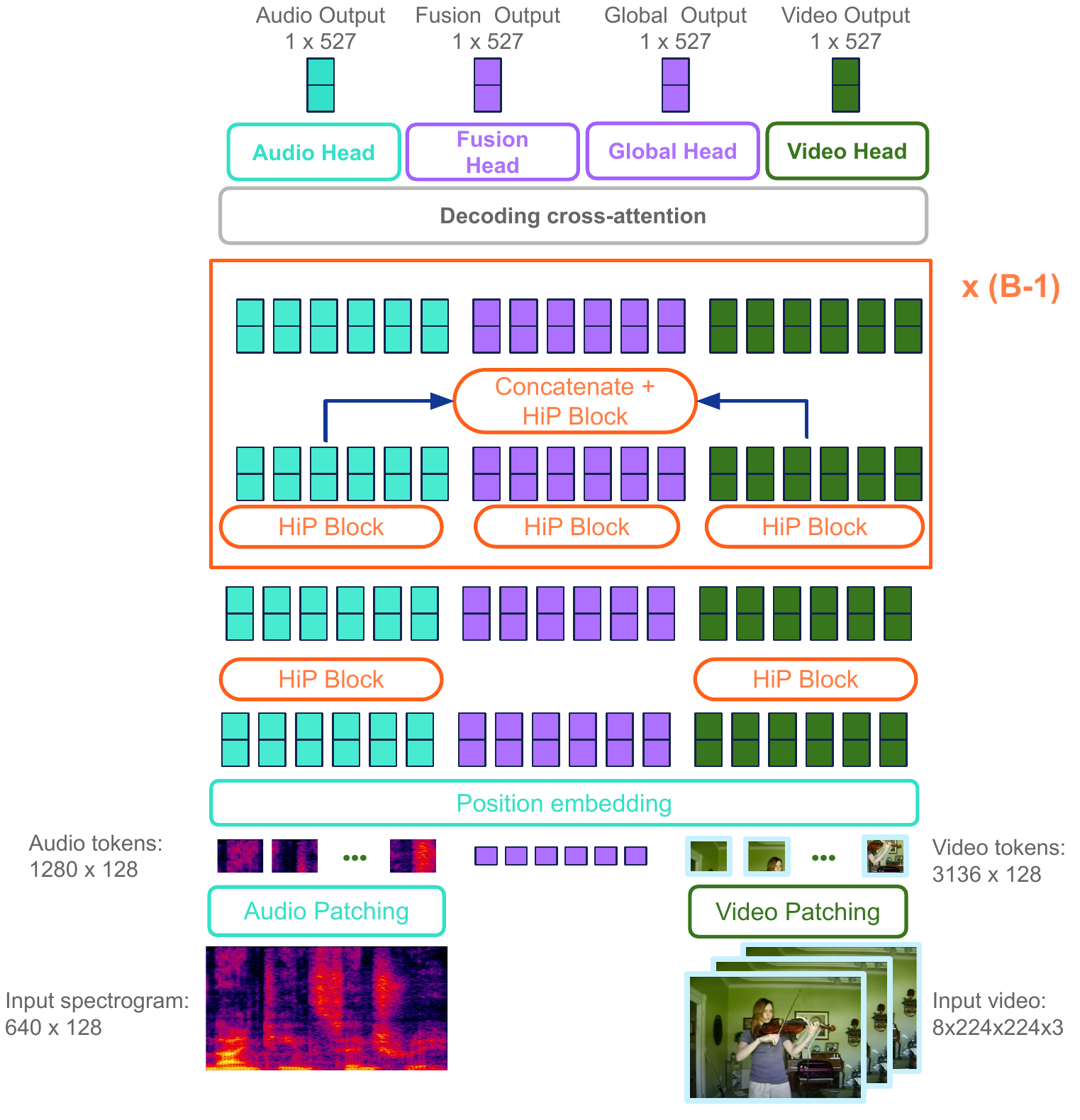}
		\caption{\textbf{\method-HiP}: The input to our model are video frames and audio spectrograms. Each of those inputs is patched using a sequence of 3D-convolutions and projection to input dimension $D=256$ . Next, each modality and the fusion tokens are processed through independent HiP blocks (which share weights). The architecture blocks the information from flowing towards the unimodal hidden representation, while still allowing the fusion representation to access all modalities. By doing this, we ensure that the video and audio representations have gated access to (i.e.\ depend on) only the video and audio inputs respectively. To produce the outputs, following Perceiver IO~\cite{perceiverIO}, we learn a set of queries that cross-attend to the unimodal and multi-modal representation. We use masking at the decoding stage to make sure we can produce unimodal outputs as well as multimodal outputs. By doing this, we can train \method using a self-supervised loss which requires unimodal representations. 
		}
		\label{fig:hip_figure}
		\vspace*{-0.05cm}
	\end{figure*}

In order to create \method-HiP, we use HiP building blocks for each modality and to process the multi-modal stream. Specifically, those blocks group the inputs into smaller sets, cross-attend using learned features and finally apply self-attention layers to the outputs of the cross attention operation (see~\cite{carreira2022hierarchical} for more details). Figure~\ref{fig:hip_figure} shows the \method-HiP architecture. Each modality is processed by a HiP model. Differently to the self-attention operation used \method-ViT and \method-Swin, in \method-HiP the multimodal tokens are processed by a HiP model. However, we skip the initial block of HiP for the fusion latents as these operations would only be applied to learned embeddings. After the three blocks are processed, we concatenate all of them together in order to create the input for the next level of the fusion stream, which splits the input in groups and cross attend to them all together. The three HiP models are sharing the same weights and architecture. We start with $6$ embedding vectors for the fusion stream, drop layers with probability $0.1$. The number of self-attend per blocks are $(2, 2, 2, 12, 2)$, the number of groups per block is $(32, 4, 1, 1, 1)$, the number of latent vectors per group in each block is $(128, 256, 256, 256, 256)$, the number of channels per block is $(256, 256, 512, 512, 1024)$, the number of heads per block is $(8, 8, 16, 16, 32)$. Finally, we using widening factor of $4$ in self-attention.

\noindent \textbf{Training speed}: For most of the experiments we use $128$ TPU-v3 cores. Here, we compare the speed of the three presented \method architectures. Under similar conditions ($8$ frames at $3.12$ FPS) and using batch size $512$, \method-ViT ($98$M parameters) has a speed of $3.2$ steps per second, \method-Swin ($161$M parameters) has a speed of $4.2$ steps per second and \method-HiP ($136$M parameters) has a speed of $5.4$ steps per second, where each step is a forward and backward pass. For \method-ViT, when we use separate weight per each stream ($267$M parameters), the speed drops to $2.8$sps. \method-HiP and \method-Swin are clearly faster models than ViT. However, as reported in Table 2 in the paper, when fine-tunning, \method-ViT still performs the best of the three architectures.  

\section{Implementation and training details}
\label{sec:training_details}
\subsection*{Audio and video pre-processing}
The inputs to our model are video and audio. The audio and video data are synced and cover the same time span. 

\vspace{2mm} \noindent\textbf{Video augmentation}: The input videos consist of $8$ frames of size $224 \times 224$. When training on AudioSet from scratch or fine-tuning in Kinetics, we sample videos at $3.12$FPS. When training in YouTube-8M, ACAV-100M or fine-tunning in VGGSound we sample at $1.25$FPS. Finally, when fine-tuning on AudioSet-500k or training \method-HiP in ACAV-100M, we sample at $1$FPS. During training, we use random cropping as well as color augmentation in the visual input. For random cropping, we sample a random bounding box which covers at least $30\%$ of the frame and the whole frame at maximum, with a random aspect ratio in the range $(0.9,1.33)$. We apply color augmentation with probability $0.8$, color randomisation on saturation and contrast $(0.6,1.4)$, brightness (max\_delta=32/255) and hue with (max\_delta=0.2). 

\vspace{2mm} \noindent\textbf{Audio augmentation}: The audio input consist of $2.56$s for training from scratch on Audioset and fine-tuning in Kinetics, and $6.4$s when pre-training on YouTube-8M and ACAV-100M or finetuning in ESC-50 and VGGSound and $8s$ when fine-tuning in AudioSet or training \method-HiP in ACAV-100M. Audio is sampled at $48kHz$. We use Log-Mel spectrograms as inputs to our model using 128 bins with Hanning windows of length $1200$ samples and stride $480$. To augment the audio, we use SpecAugment~\cite{park19specaug} and frequency jittering where we shift the frequency by an integer sampled on the range $(-10,10)$. When fine-tuning ESC-50, we use the same input length as in pre-training ($6.4$s). As the input length is larger than ESC-50 samples, we loop over the samples.

\vspace{2mm} \noindent\textbf{Evaluation details}:
For evaluation, we use equally spaced $8$ frame clips with the same stride as during training. Furthermore, we take $3$ crops for each of those clips and average the resulting predictions for multi-crop evaluation. When evaluating ESC-50, we feed a single clip as it covers the whole length of the signal. We report performance in the test set for AudioSet and VGGSound and validation set for Kinetics-400. For ESC-50, we average the performance of the $5$ splits. 

\vspace{2mm} \noindent\textbf{Optimisation details}:
Details about optimizers and hyperparameters for models used in the paper can be found in Table~\ref{tab:hyper}. For AudioSet from scratch we select the highest learning rate (using $0.0003$ as maximum) that does not lead to collapse. When fine-tuning, we choose the learning rate by evaluating without augmentation. In ESC-50, we use the first split to select the learning rate. We train all our models with batch size $512$ and we use learning rate scaling with factor $\frac{\textit{batch size}}{256}$. We train all models for $50$ epochs except for the ACAV-100M models which we train for $10$ epochs. For the ablations \textit{Bottleneck} and \textit{Vanilla}, we train for $25$ epochs to prevent overfitting. For fine-tuning in AudioSet-500k, we use label smoothing of $0.15$ and modality mixup~\cite{zhang2017mixup}. Different from \cite{mbt}, we need to provide supervision not only for the multimodal output but also for the unimodal output. This means the mixup procedure can be performed in many different ways. We find sampling a single mixup value from a $\beta(0.3,0.3)$ the best configuration to apply mixup.

\vspace{2mm} \noindent\textbf{Contrastive learning}:
In order to pre-train \method using the audio-visual contrastive loss, we define two projectors for the audio and video ouputs of the model, with different weights. When training the model with the complete contrastive loss involving the fusion weights, we create a third projector for the fusion output. Those projectors consist of an MLP using hidden dimensionality of $512$. We use temperature $\tau=0.08$ for the training loss.

\section{Fusion position}
\label{arch:choices}
In this section, we extend the study presented in the main paper to study the effect of introducing the fusion tokens starting at a certain level of the network, inspired by~\cite{mbt}. For these experiments, we trained our models on AudioSet-2M for $25$ epochs and report performance using the standard evaluation protocol as described in Section~\ref{sec:training_details}.
	\begin{table}[t]
		\centering
		\caption{\small {\bf Modality fusion position.} We report the performance of our models trained on audio-visual data when the multi-modal fusion is done at different layers. 
        }
			\begin{tabular}{c|ccc} \toprule
			Fusion level  & Audio & Video & Audio+Video \\
			 \hline
			 0& 37.6  & 20.8  &44.2 \\
			 3 & 38.0& 21.4 & 44.6\\
			 6 &37.5 &20.3  &44.2 \\
			 9 &37.6 &20.9  &44.4 \\

			\bottomrule
		\end{tabular}
		\label{tab:layer_fusion}
	\end{table}

In Table~\ref{tab:layer_fusion} we report the performance of our model when introducing the fusion stream at a different layer of the network. In previous layers, the fusion tokens are not used. Interestingly, the position of where to introduce the fusion layer does not seem critical to the overall performance of the model. We attribute this to the fact that Zorro keeps its unimodal streams untouched, preventing the full representation from overfitting to the most informative modality for a given task. In order to align with standard architectures, we choose to use our fusion layers from the beginning of the model.

	\begin{table*}[t]
		\centering
		\caption{\small {\bf Hyperparameters.} For reproducibility, in this table we report the hyperparameters used for each model in the paper. 
        }
			\begin{tabular}{cccc|ccc} \toprule
			Model  & Sup. & Dataset & Scratch/Fine-tunning & Optimizer & Learning Rate & Weight Decay \\
			 \hline
			 \method-ViT  & Supervised &  AS & Scratch   & Adam & $0.0003$ & $10^{-6}$\\
			 Two Streams  & Supervised &  AS & Scratch   & Adam & $0.0001$ & $10^{-6}$ \\
			 Vanilla Fusion & Supervised &  AS & Scratch   & Adam & $0.0001$ & $10^{-6}$ \\
			 Bottleneck Fusion  & Supervised &  AS & Scratch   & Adam & $0.0001$ & $10^{-6}$ \\
			 \method-ViT  & Contrastive &  AS & Scratch   & Adam & $0.00005$ & $10^{-6}$ \\
			 Two Streams  & Contrastive &  AS & Scratch   & Adam & $0.00005$ & $10^{-6}$\\
			 \method-Swin  & Supervised &  AS & Scratch   & Adam & $0.0001$ & $10^{-6}$\\
			 \method-HiP  & Supervised &  AS & Scratch   & Adam & $0.0001$ & $10^{-6}$\\
			 \method-Swin  & Contrastive &  AS & Scratch   & Adam & $0.0001$ & $10^{-6}$\\
			 \method-HiP  & Contrastive &  AS & Scratch   & Adam & $0.00001$ & $10^{-6}$ \\
			 \method-ViT  & Supervised &  YT8M & Scratch   & Adam & $0.00008$ & $10^{-6}$\\
			 \method-ViT  & Contrastive &  ACAV & Scratch   & Adam & $0.00005$ & $10^{-6}$\\
			 \method-Swin  & Contrastive &  ACAV & Scratch   & Adam & $0.00005$ & $10^{-6}$ \\
			 \method-HiP  & Contrastive &  ACAV & Scratch   & Adam & $0.0001$ & $10^{-6}$ \\
			 \method-ViT  & Supervised &  AS-500k & FT (ACAV-100M)   & SGD & $0.08$ & $10^{-6}$\\
			 \method-ViT  & Supervised &  AS-500k & FT (YT-8M)  & SGD & $0.06$ & $0$ \\
			 \method-ViT  & Supervised &  AS-500k & FT (IN-21k)  & SGD & $0.1$ & $0$ \\
			 \method-Swin  & Supervised &  AS-500k & FT (ACAV-100M)   & SGD & $0.05$ & $0$ \\
			 \method-HiP  & Supervised &  AS-500k & FT (ACAV-100M)  & Adam & $0.0001$ & $0$ \\
			 \method-ViT  & Supervised &  VGGSound & FT (ACAV-100M)   & SGD & $0.01$ & $0$ \\
			 \method-ViT  & Supervised &  VGGSound & FT (YT-8M)  & SGD & $0.01$ & $0$\\
			 \method-ViT  & Supervised &  VGGSound & FT (IN-21k)  & SGD & $0.08$ & $0$\\
			 \method-Swin  & Supervised &  VGGSound & FT (ACAV-100M)   & SGD & $0.05$ & $0$ \\
			 \method-HiP  & Supervised &  VGGSound & FT (ACAV-100M)  & Adam & $0.0001$ & $0$\\

			 \method-ViT  & Supervised &  K400 (V+A) & FT (ACAV-100M)   & SGD & $0.05$ & $0$\\
			 \method-ViT  & Supervised &  K400 (V+A) & FT (YT-8M)  & SGD & $0.07$ &  $0$ \\
			 \method-ViT  & Supervised &  K400 (V+A) & FT (IN-21k)  & SGD & $0.08$ &  $0$ \\

			 \method-Swin  & Supervised &  K400 (V+A) & FT (ACAV-100M)   & SGD & $0.05$ &  $0$\\
			 \method-HiP  & Supervised &  K400 (V+A) & FT (ACAV-100M)  & Adam & $0.0001$ &  $0$\\
			 \method-ViT  & Supervised &  K400 (V) & FT (ACAV-100M)   & SGD & $0.08$ &  $0$\\
			 \method-ViT  & Supervised &  K400 (V)  & FT (YT-8M)  & SGD & $0.08$ & $0$ \\
			 \method-ViT  & Supervised &  K400 (V)  & FT (IN-21k)  & SGD & $0.08$ &  $0$\\
			 \method-Swin  & Supervised &  K400 (V)  & FT (ACAV-100M)   & SGD & $0.05$ &  $0$\\
			 \method-HiP  & Supervised &  K400 (V)  & FT (ACAV-100M)  & Adam & $0.0001$ &  $0$ \\
			 \method-ViT  & Supervised &  ESC-50 & FT (ACAV-100M)   & Adam & $ 0.0009$ &  $0.001$\\
			 \method-ViT  & Supervised &  ESC-50 & FT (YT-8M)  & Adam & $ 0.0009$ & $0.001$ \\
			 \method-ViT  & Supervised &  ESC-50  & FT (IN-21k)  & Adam & $ 0.0009$ &  $0.001$\\
			 \method-Swin  & Supervised &  ESC-50  & FT (ACAV-100M)   & Adam & $0.0007$ &  $0$\\
			 \method-HiP  & Supervised &  ESC-50  & FT (ACAV-100M)  & Adam & $0.0003$ &  $0$ \\

			\bottomrule
		\end{tabular}
		\label{tab:hyper}
	\end{table*}

\end{document}